\documentclass[11pt]{article}
\usepackage[T1]{fontenc}
\usepackage[utf8]{inputenc}
\usepackage{amsmath}
\usepackage{amsfonts}
\usepackage{graphicx}
\usepackage{float}
\usepackage{subcaption}
\usepackage{booktabs}
\usepackage{hyperref}
\usepackage{cite}
\usepackage{url}
\usepackage[margin=1in]{geometry}
\usepackage{fancyhdr}
\usepackage{lipsum}
\usepackage{algorithm}
\usepackage{algpseudocode}

\title{TextDiffuser-RL: Efficient and Robust Text Layout Optimization for High-Fidelity Text-to-Image Synthesis}

\author{
\begin{tabular}{c}
Kazi Mahathir Rahman$^{1}$, Showrin Rahman$^{2}$, Sharmin Sultana Srishty$^{3}$ \\
$^{1}$BRAC University, Dhaka, Bangladesh \\
\texttt{kazi.mahathir.rahman@g.bracu.ac.bd}, \texttt{showrin.rahman@g.bracu.ac.bd}, \\
\texttt{sharmin.sultana.srishty@g.bracu.ac.bd}
\end{tabular}}

\begin{document}
\maketitle

\begin{abstract}
Text-embedded image generation plays a critical role in industries such as graphic design, advertising, and digital content creation. Text-to-Image generation methods leveraging diffusion models, such as TextDiffuser-2\cite{textdiffuser2}, have demonstrated promising results in producing images with embedded text. TextDiffuser-2 effectively generates bounding box layouts that guide the rendering of visual text, achieving high fidelity and coherence. However, existing approaches often rely on resource-intensive processes and are limited in their ability to run efficiently on both CPU and GPU platforms. To address these challenges, we propose a novel two-stage pipeline that integrates reinforcement learning (RL) for rapid and optimized text layout generation with a diffusion-based image synthesis model. Our RL-based approach significantly accelerates the bounding box prediction step while reducing overlaps, allowing the system to run efficiently on both CPUs and GPUs. Extensive evaluations demonstrate that our framework achieves comparable performance to TextDiffuser-2 in terms of text placement and image synthesis, while offering markedly faster runtime and increased flexibility. Our method produces high-quality images comparable to TextDiffuser-2, while being \textbf{42.29$\times$ faster} and requiring only \textbf{2MB of CPU RAM} for inference unlike TextDiffuser-2’s M1 model, which is not executable on CPU-only systems.
\end{abstract}


\section{Introduction}
Text-embedded image generation is a cross-disciplinary task at the intersection of computer vision and natural language processing. It involves generating images that embed textual content in a visually coherent and semantically accurate manner, with applications in poster design, advertisements, educational media, and visual storytelling. Controlling both the content and layout of text is critical, as spelling or positioning errors can reduce the effectiveness of the output.

Diffusion-based models, such as TextDiffuser-2~\cite{textdiffuser2}, have significantly advanced text-to-image generation by introducing a two-stage pipeline for high-quality text rendering. However, these models often suffer from layout collisions, overlapping boxes, and spelling errors, along with high computational costs that hinder real-time applications. Moreover, SA-OcrPaint~\cite{saocrpaint} attempts to mitigate these issues using OCR-guided layout refinement and inpainting. While it improves fidelity, its iterative process is inefficient and lacks adaptability to diverse layout scenarios.

To overcome these limitations, we propose a reinforcement learning (RL)-based approach for layout planning. RL offers a flexible framework for optimizing layouts using feedback from OCR accuracy and CLIP-based perceptual similarity, allowing policies that generalize across diverse prompts. Our framework, \textbf{TextDiffuser-RL}, introduces a lightweight RL-based planning module that decouples layout generation from image synthesis. In the first stage, a custom environment called \textit{GlyphEnv} trains an RL agent to generate non-overlapping, spatially aware bounding boxes. The second stage invokes a diffusion model for image generation using these layouts, improving efficiency and quality. We evaluate our method using the MARIOEval~\cite{textdiffuser} dataset, a standardized benchmark derived from MARIO-10M, designed to test layout-aware text rendering. It offers high-quality caption-OCR pairs, making it ideal for evaluating both structural and semantic fidelity in generated layouts.

\begin{figure*}[!hbt]
    \centering
    \includegraphics[width=\linewidth]{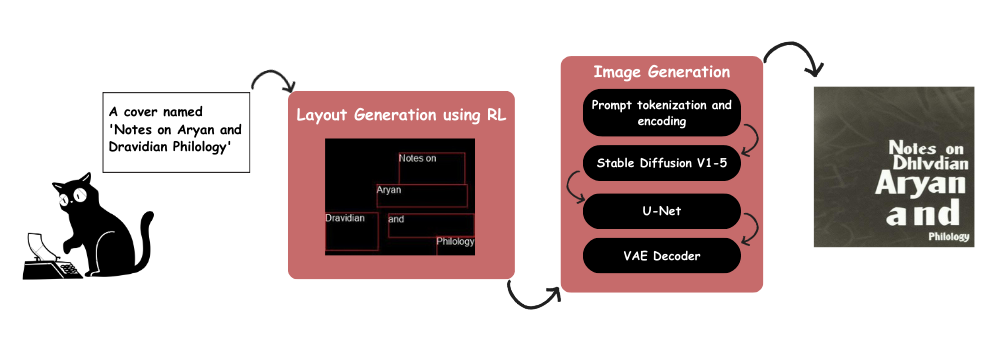}
    \caption{Overview of the model architecture, showing the diffusion-based synthesis process with GlyphEnv optimizing non-overlapping bounding box generation, layout planning, and quality assessment based on OCR accuracy and spatial alignment.}
    \label{fig:model_architecture}
\end{figure*}

In the second stage, the generated layout is passed to a modified Stable Diffusion model equipped with a VEA decoder, which functions similarly to TextDiffuser-2. This architecture ensures that the diffusion model receives well-structured input, thereby improving text rendering, reducing spelling mistakes, and minimizing layout collisions. Experimental results on MARIOEval demonstrate that our method is 42.29$\times$ faster and uses only 2 MB of memory, while maintaining competitive OCR accuracy and CLIPScore. These results confirm the effectiveness of combining layout-aware reinforcement learning with generative diffusion models for text-embedded image synthesis.

In this paper, we propose a novel framework, \textbf{TextDiffuser-RL}, which advances the field of text-embedded image generation by introducing the following key contributions:
\begin{itemize}
    \item We propose \textbf{TextDiffuser-RL}, the first framework to integrate reinforcement learning-based layout optimization with diffusion-based text-embedded image generation.
    \item We introduce \textbf{GlyphEnv}, a novel RL environment tailored for generating non-overlapping and visually coherent text layouts using PPO algorithms.
    \item We show that TextDiffuser-RL significantly reduces runtime and memory usage while maintaining text accuracy, as validated on the MARIOEval benchmark.
\end{itemize}

\section{Related Work}
Generating images with embedded text is still a difficult task, especially when trying to show text clearly and correctly inside images. The MARIO-Eval benchmark \cite{textdiffuser} has become an important standard for measuring progress in this field. Recently, several studies have made strong improvements by developing better methods for both arranging text layouts and creating images.

One such method, UDiffText \cite{zhao2023udifftext}, is a framework that uses a pre-trained diffusion model to generate images containing text. It features a lightweight character-level text encoder and fine-tunes the diffusion model using large datasets. This approach achieves high accuracy when synthesizing text within different images, and it performs well in tasks like text-centered image creation and scene text editing.

Another approach, Glyph-ByT5 \cite{liu2024glyphbyt5}, introduces a custom text encoder that fine-tunes the character-aware ByT5 model using paired glyph and text data. This encoder aligns closely with how text looks visually, greatly improving how accurately text is rendered. When used together with the SDXL model, Glyph-ByT5 improved text rendering accuracy on design image benchmarks from less than 20 percent to nearly 90 percent, showing its power in handling complex text rendering challenges.

TextDiffuser \cite{textdiffuser2} was one of the first methods to use a two-stage diffusion model specially designed for images with lots of text. Its process uses a transformer-based layout planner to create bounding boxes for keywords, followed by a diffusion U-Net that generates the image based on this layout. TextDiffuser also introduced the MARIO-10M dataset and the MARIO-Eval benchmark, establishing a strong baseline in text rendering measures like OCR accuracy and F1 score. However, the model still had room to improve in spelling accuracy and how well the layout was organized.

Building on this, TextDiffuser-2 \cite{textdiffuser2} improves text layout and image generation by using large language models (LLMs). These fine-tuned LLMs create and encode text layouts at the line level, which allows for more complex and varied text arrangements. TextDiffuser-2 showed better results than the original model, especially on the MARIO-Eval benchmark, showing how LLMs help produce more natural text placement and clearer rendering.

The ARTIST framework \cite{artist2023} introduced a separate structure that splits text modeling and image creation into two parts. It uses a pretrained text diffusion model to guide a visual diffusion model, leading to big improvements in text accuracy and alignment. ARTIST notably outperformed earlier methods on MARIO-Eval by a large margin in OCR accuracy and F1 score, showing how separating learning tasks and adding fine control can boost text-to-image synthesis.

Another approach is SA-OcrPaint \cite{saocrpaint}, which focuses on improving text in images without needing to retrain models. It uses simulated annealing to optimize the layout of glyphs and OCR-guided inpainting to fix spelling mistakes. This method effectively increases text accuracy in generated images, showing strong improvements on the MARIO-Eval dataset. It offers a practical and hardware-friendly way to refine diffusion model outputs.

In "Large-scale Reinforcement Learning for Diffusion Models" \cite{zhang2024large}, the authors propose a scalable reinforcement learning algorithm that improves diffusion models by using various reward functions, including human preferences and image composition. This method greatly enhances how well generated images match human expectations, improving both the composition and variety of images. Another work, "DPOK: Reinforcement Learning for Fine-tuning Text-to-Image Diffusion Models" \cite{zhang2023dpok}, presents an online reinforcement learning technique to fine-tune pre-trained text-to-image diffusion models. Combined with policy optimization and KL regularization, this approach improves image alignment and overall quality compared to traditional supervised fine-tuning.

The study "Reinforcement Learning with Human Feedback for Generative Models" \cite{ouyang2023reinforcement} investigates using reinforcement learning guided by human feedback to fine-tune generative models. By using classifier guidance, this method builds on text-image alignment techniques. This allows more precise control over visual elements based on the text descriptions.

\section{Methodology}
To tackle the challenges in text-to-image generation, we introduce a two-stage process. In the first stage, reinforcement learning (RL) is used to optimize bounding box layouts by reducing overlaps and perfectly placing the bounding box. In the second stage, a diffusion model takes these optimized layouts as guidelines to create high-quality images. The detailed steps are explained below.

\begin{figure}[H]
    \centering
    \includegraphics[width=\linewidth, height=8cm]{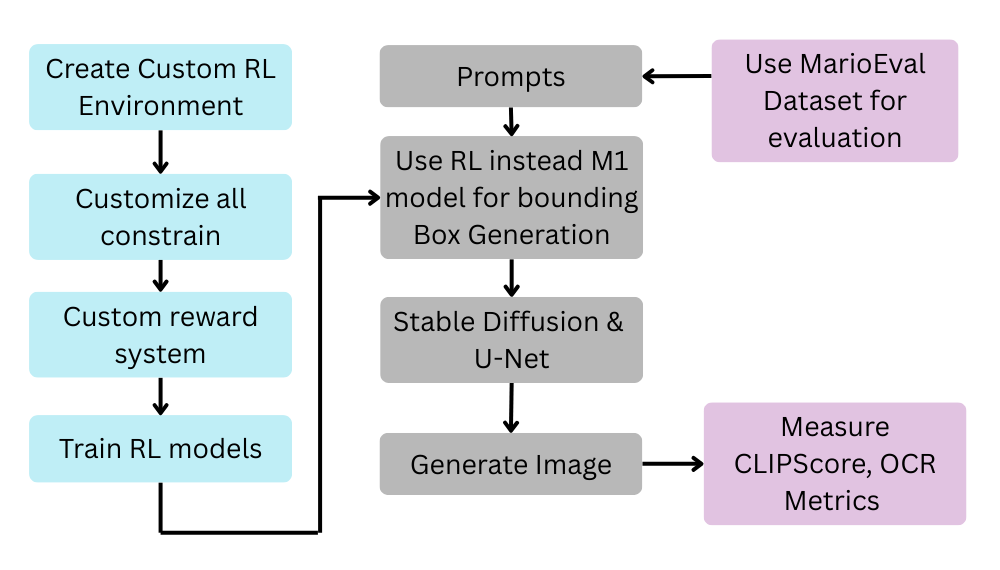}
    \caption{Two-stage Workflow}
    \label{fig: Two-stage pipeline}
\end{figure}

\subsection{Environment Architecture}
To support layout optimization in text-to-image generation, we created a custom reinforcement learning environment called \texttt{GlyphEnv}. This was built using the Gymnasium framework. This environment is designed for placing bounding boxes in an organized way within a fixed 2D space. It takes continuous actions and evaluates each state based on how well it meets some constrains and layout quality. The main goal is to generate bounding boxes that avoid overlapping with size and shape requirements. This helps ensure the final images are semantically clear and well-structured in later steps of the process.

\begin{algorithm}
\caption{GlyphEnv Environment Pseudocode}
\begin{algorithmic}[1]
\State \textbf{Initialize:} window\_size, min\_overlap, num\_rectan, min\_area, w\_min, h\_min
\State
\Function{GenerateInitialState}{}
    \For{$i = 1$ to num\_rectan}
        \State Randomly sample $(x_1, y_1)$
        \State Compute width $w$ and height $h$ such that $w \times h = \text{min\_area}$,
        \State \quad with constraints inside window
        \State Set $(x_2, y_2) \gets (x_1 + w, y_1 + h)$
        \State Store rectangle $i \gets [x_1, y_1, x_2, y_2]$
    \EndFor
    \State \Return list of rectangles
\EndFunction
\State

\Function{Reset}{}
    \State state $\gets$ \Call{GenerateInitialState}{}
    \State \Return state
\EndFunction
\State

\Function{Step}{action}
    \For{each rectangle $i$ in state}
        \State Apply action deltas $(\Delta x_1, \Delta y_1, \Delta x_2, \Delta y_2)$
        \State Clip coordinates to stay inside window
        \State Recalculate width and height
        \State Update rectangle $i$ with new coordinates
    \EndFor

    \State overlap\_penalty $\gets$ \Call{CalculateOverlap}{state}
    \If{overlap\_penalty $<$ min\_overlap}
        \State reward $\gets$ positive reward proportional to $(1 - \frac{\text{overlap\_penalty}}{\text{min\_overlap}})$
        \State done $\gets$ \textbf{true}
    \Else
        \State reward $\gets$ negative penalty proportional to $\frac{(\text{overlap\_penalty} - \text{min\_overlap})}{(1 - \text{min\_overlap})}$
        \State done $\gets$ \textbf{false}
    \EndIf

    \State \Return updated state, reward, done, info
\EndFunction
\State

\Function{CalculateOverlap}{rectangles}
    \State overlap\_sum $\gets 0$
    \For{each unique pair $(i, j)$ of rectangles}
        \State overlap\_sum += \Call{CalculateIoU}{$i$, $j$}
    \EndFor
    \State \Return overlap\_sum
\EndFunction
\State

\Function{CalculateIoU}{box1, box2}
    \State Compute the intersection area of box1 and box2
    \State Compute the union area of box1 and box2
    \State \Return intersection / union \textbf{if} union $>$ 0 \textbf{else} 0
\EndFunction

\end{algorithmic}
\end{algorithm}

\subsection{Constraint Modeling}

The design of \texttt{GlyphEnv} includes several geometric and spatial rules to ensure the generated layouts are useful:
\begin{itemize}
    \item \textbf{Minimum Area}: Each bounding box must cover at least a certain amount of area. This helps keep the text inside readable and visually noticeable. The box size is also constrained so it doesn’t become too small to fit content or too large to dominate the layout.
    \item \textbf{Minimum Dimensions}: Boxes must meet set minimum width ($w_{\min}$) and height ($h_{\min}$) values to avoid shapes that are too narrow or small for proper text display. This ensures that text does not get compressed into overly slim fonts, which affects readability.
    \item \textbf{Boundary Confinement}: All box coordinates must stay within a fixed square area of size \ensuremath{\text{window\_size} \times \text{window\_size}}. All boxes must lie entirely within a certain window to prevent content from being cut off or positioned outside of the image canvas.
\end{itemize}

These rules are built into both the starting layout and the way states are updated during training. This ensures these constraints are always followed.

\subsection{Action and Observation Spaces}
The environment uses a continuous action space. For each of the $N$ bounding boxes, the agent outputs a 4-dimensional vector representing small changes: $(\Delta x_1, \Delta y_1, \Delta x_2, \Delta y_2)$. These values adjust the top-left and bottom-right corners of each box. The overall action space has the shape $(N, 4)$ with each delta limited to the range $[-1, 1]$.

The observation space is a real-valued matrix with shape $(N, 4)$, where each row holds the coordinates $(x_1, y_1, x_2, y_2)$ of a bounding box. These values are clipped to stay within the canvas boundaries and are updated based on the agent’s actions.

\subsection{Reward Formulation}
To help the agent learn effective layout configurations, we design a reward function that penalizes overlapping text regions and rewards layouts where the boxes are well separated. The reward is based on the total Intersection-over-Union (IoU) calculated across all pairs of bounding boxes. A lower overlap leads to a higher reward and has cleaner and more readable layouts. Formally, the reward $R$ is defined as:

\begin{equation}
R = \begin{cases}
\textstyle 10\left(1 - \frac{o}{m}\right), & \text{if } o < m \\
\textstyle -10\left(\frac{o - m}{1 - m}\right), & \text{otherwise}
\end{cases}
\label{eq:reward}
\end{equation}

where \( o = \text{Overlap} \), \( m = \text{min\_overlap} \).The formulation encourages rapid convergence to low-overlap states, which are crucial for downstream image coherence and text readability. The reward sharply penalizes high overlaps while rewarding low overlaps, which drives the agent to produce better layouts quickly. It is normalized and piecewise design ensures stable training and effective learning across complex constraints for layout generation.

\subsection{Overlap Metric}
A key aspect of the environment is its overlap evaluation metric. The pairwise IoU \cite{jaccard1901} between any two boxes $i$ and $j$ is computed as:

\begin{equation}
\text{IoU}_{i,j} = \frac{|B_i \cap B_j|}{|B_i \cup B_j|}
\label{eq:iou}
\end{equation}

where $|B_i \cap B_j|$ and $|B_i \cup B_j|$ represent the intersection and union areas of boxes $i$ and $j$, respectively. The cumulative overlap metric used for reward calculation is obtained by summing IoU values across all unique box pairs:

\begin{equation}
\text{Total Overlap} = \sum_{i=1}^{N} \sum_{j=i+1}^{N} \text{IoU}_{i,j}
\label{eq:total_overlap}
\end{equation}

This metric captures the frequency and extent of overlaps, providing a strong signal for training the agent to learn spatially aware layout strategies. By combining these elements, the environment provides a clear and effective setup for layout optimization. This makes for promising applications in multimodal content generation systems.

\subsection{Training a Reinforcement Learning Algorithm with a Custom Environment}

We trained three reinforcement learning algorithms—Proximal Policy Optimization (PPO) \cite{schulman2017proximal}, Deep Deterministic Policy Gradient (DDPG) \cite{ddpg}, and Soft Actor-Critic (SAC) \cite{haarnoja2018soft}—in our custom continuous control environment, \texttt{GlyphEnv}. This environment is built for optimizing text layouts, where the agent’s actions directly control the positions and sizes of bounding boxes.

Training was done efficiently using only a CPU, with low requirements for time and memory. The experiments were run on a standard Google Colab setup with a 2-core Intel Xeon CPU and 12–16 GB of RAM. This shows that our method was trained without needing powerful or expensive GPU resources.

\begin{figure}[h]
    \centering
    \begin{subfigure}[b]{0.48\linewidth}
        \centering
        \includegraphics[width=\linewidth, height=4cm]{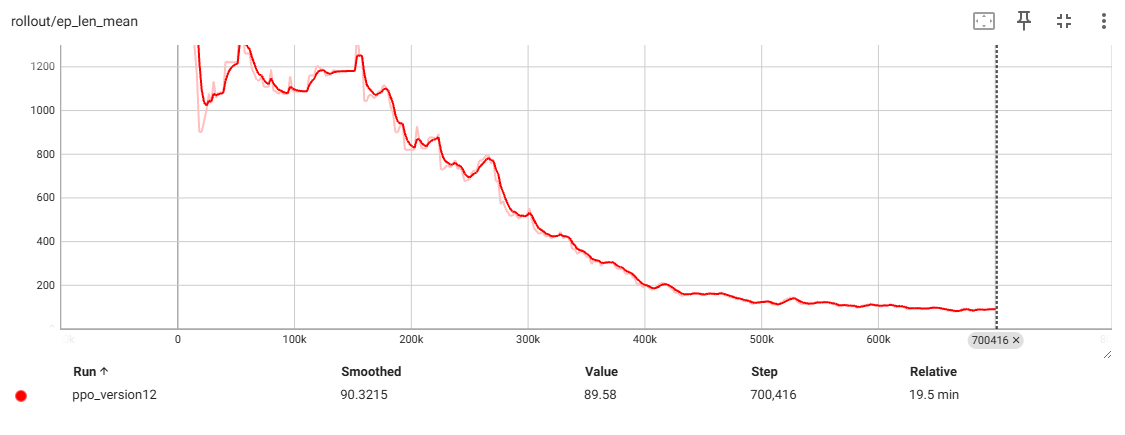}
        \label{fig:length_curve}
    \end{subfigure}
    \hfill
    \begin{subfigure}[b]{0.48\linewidth}
        \centering
        \includegraphics[width=\linewidth, height=4cm]{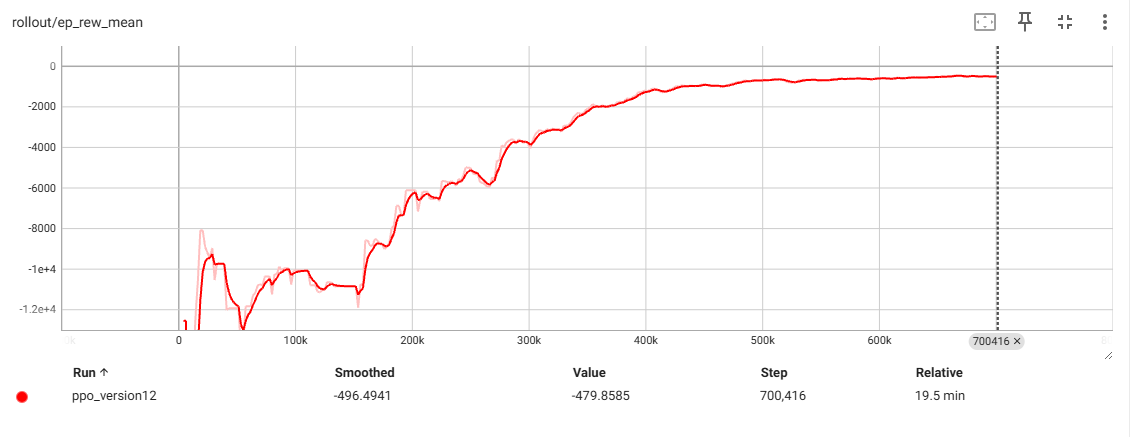}
        \label{fig:reward_curve}
    \end{subfigure}
    \caption{Comparison of PPO training performance across episodes. Episode length comparison across PPO, showing stable and shorter episodes over time, indicating faster task completion. Mean episode reward curve consistently improves over time.}
    \label{fig:ppo_results}
\end{figure}

Among the three methods, PPO showed the best training stability and performance, as seen in the average episode reward and episode length plots (see Figures~\ref{fig:ppo_results}). PPO’s clipped objective allows it to learn strong and reliable policies. Even when rewards are sparse or change over time, conditions that are common in layout generation tasks. On the other hand, SAC and DDPG had difficulty converging because of instability from their off-policy updates and the challenges posed by the tightly constrained action space.

\begin{figure}[H]
    \centering
    \includegraphics[width=0.3\linewidth]{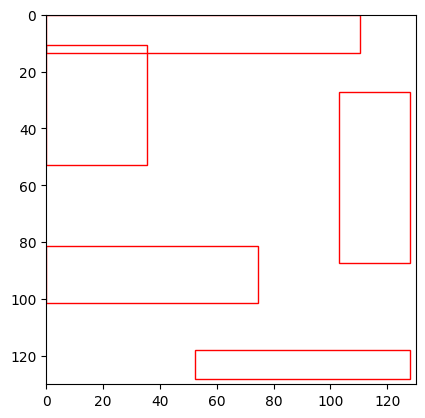}
    \includegraphics[width=0.3\linewidth]{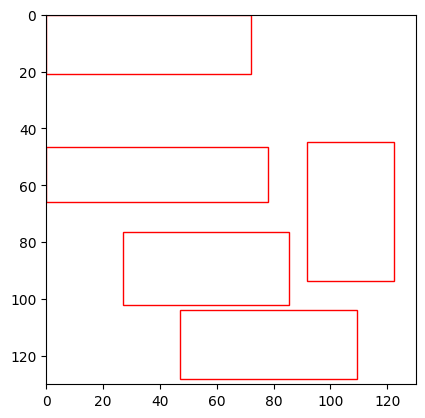}
    \includegraphics[width=0.3\linewidth]{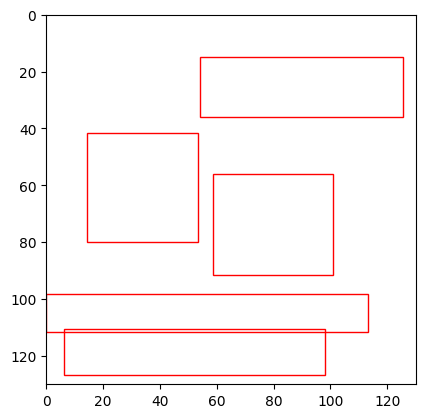}
    \caption{RL generated bounding box}
    \label{fig:rl_generated_bounding_box}
\end{figure}

\subsection{TextDiffuser-RL Model for Image Generation}
We replace the original bounding box generation module with boxes predicted by reinforcement learning (RL) agents in our TextDiffuser-RL model. These agents produce optimized layouts that reduce overlap and improve text positioning. The pipeline starts with the \texttt{tokenizer}, which converts the input prompt into tokens. These tokens are passed to the \texttt{text\_encoder} to generate embeddings used in the diffusion process. The bounding boxes from the RL agent act as spatial constraints that guide the \texttt{ddpm\_scheduler}~\cite{ho2020denoising}, which controls the denoising steps during image generation. The \texttt{U-Net} then refines the latent representation by predicting and removing noise. Finally, the \texttt{vae} decoder turns the latent tensor into the final image. By integrating RL-based layouts into this process, the model achieves faster and more accurate control over text placement in the generated images.

\section{Experiments}

\subsection{Reinforcement Learning Algorithm selection}
To evaluate how well each reinforcement learning algorithm performs in optimizing text layouts, we tested PPO, SAC, and DDPG. However, three main metrics: mean reward, standard deviation of rewards, and average IoU overlap between bounding boxes were used for comparing the model's convergence ability. Among the three, PPO delivered the best results, achieving a mean reward of 579.54 and an average IoU overlap of 0.26. This indicates that PPO consistently produced layouts with low overlap, making it the most effective for this task.

\begin{table}[H]
\centering
\footnotesize % or \scriptsize if still too wide
\begin{tabular}{|c|c|c|c|}
\hline
\textbf{Model} & \textbf{Mean Reward} & \textbf{Reward Std Dev} & \textbf{Mean IOU} \\
\hline
PPO  & \textbf{579.54}  & \textbf{2207.42} & \textbf{0.26} \\
SAC  & -4222.90         & 4576.17          & 0.61 \\
DDPG & -6200.37         & 3216.42          & 0.70 \\
\hline
\end{tabular}
\caption{Performance comparison of PPO, SAC, and DDPG over 200 episodes (1000 steps max each) in GlyphEnv }
\label{tab:rl_comparison}
\end{table}

We trained our PPO agent using a learning rate of $3 \times 10^{-4}$, a rollout horizon of 2048 steps, and a batch size of 64. We used 10 epochs per update and set the discount factor to $\gamma = 0.99$ with GAE parameter $\lambda = 0.95$. The clipping range was set to 0.2, and gradient updates were clipped at a maximum norm of 0.5.

\begin{figure}[H]
    \centering
    \includegraphics[width=0.9\linewidth, height=4cm]{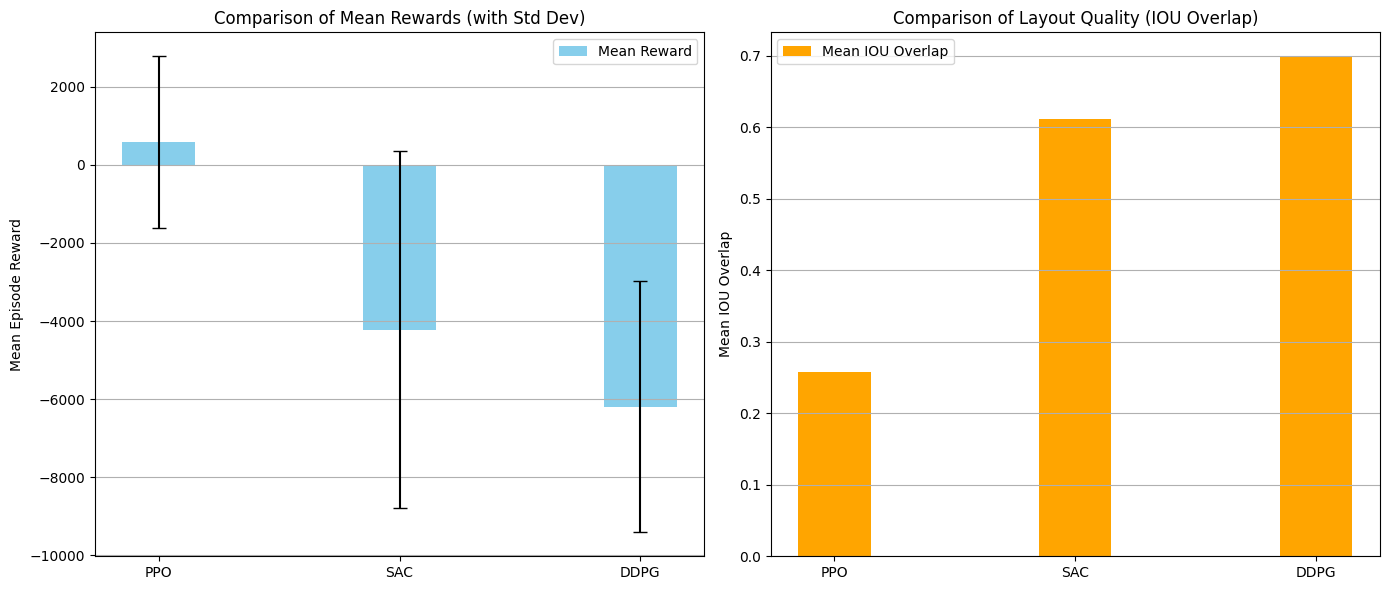}
    \caption{Visual comparison of RL model performances based on bounding box layouts}
    \label{fig:rl_comparison}
\end{figure}

\subsection{Ablation Study}

\subsubsection{Rectangular Numbers and Minimum Area Selection}

We performed an ablation study to see how changing the number of rectangles and the minimum area constraint affects the agent’s performance. As expected, increasing the number of rectangles from 4 to 7 made the layout task harder. The reward dropped drastically, and the IoU overlap increased, which shows the agent struggled to place more rectangles without much overlap. For example, with 4 rectangles, the model earned a reward of –113.47 and an IoU of 0.3389, but with 7 rectangles, the reward fell to –9198.60 and the IoU rose to 1.0010. This indicates a clear layout failure. We also tested how different minimum area constraints impacted performance. When the minimum area increased from 1300 to 2000, the agent had less freedom to place boxes. Although the IoU rose slightly, showing more overlap, the overall reward tended to decrease and reflects the difficulty of fitting larger boxes inside a fixed canvas.

\begin{table}[ht]
\centering
\begin{tabular}{|c|c|c|c|}
\hline
\multicolumn{4}{|c|}{\textbf{Effect of Number of Rectangles}} \\
\hline
\textbf{Num Rect} & \textbf{Mean Reward} & \textbf{Std} & \textbf{Mean IoU} \\
\hline
4 & -113.47 & 136.57 & 0.3389 \\
5 & -378.82 & 295.08 & 0.5416 \\
6 & -1487.96 & 1283.24 & 0.6825 \\
7 & -9198.60 & 3990.78 & 1.0010 \\
\hline
\multicolumn{4}{|c|}{\textbf{Effect of Minimum Area}} \\
\hline
\textbf{Min Area} & \textbf{Mean Reward} & \textbf{Std} & \textbf{Mean IoU} \\
\hline
1300 & -454.98 & 468.68 & 0.4971 \\
1500 & -458.24 & 441.07 & 0.5523 \\
1800 & -808.25 & 596.77 & 0.5743 \\
2000 & -612.12 & 318.18 & 0.6522 \\
\hline
\end{tabular}
\caption{Effect of different numbers of rectangles and the minimum area on model performance. This evaluation is over 200 episodes and 1000 steps per episode and window size 128$\times$128.}
\label{tab:ablation_combined}
\end{table}

We selected 5 rectangles with a minimum area of 1500 based on a balance between task complexity, visual clarity, and training stability. Using fewer boxes led to overly simplistic layouts that did not hold a sufficiently number of words. While more boxes or smaller areas introduced excessive overlap and ambiguity, hindering learning and degrading visual quality. The choice of five reasonably sized boxes also aligns with common design patterns in poster.

\subsubsection{Random Seed Robustness}
We ran our model using five different random seeds (0, 42, 123, 551, and 999) to check its stability. Despite starting from different points, the PPO agent gave fairly consistent results. The average rewards ranged from about –325 to –478, and the IoU overlap scores stayed close, between 0.52 and 0.56. The reward variations were also moderate, showing that training is stable and not heavily influenced by randomness. Overall, the model learns reliably regardless of the seed, which is important for trusting it in real-world applications.

\begin{table}[h!]
\centering
\begin{tabular}{|c|c|c|c|}
\hline
\textbf{Seed} & \textbf{Mean Reward} & \textbf{Std Dev Reward} & \textbf{Mean IoU} \\
\hline
0   & -402.08 & 363.37 & 0.5153 \\
42  & -348.03 & 335.61 & 0.5418 \\
123 & -356.70 & 253.77 & 0.5393 \\
551 & -325.36 & 214.91 & 0.5599 \\
999 & -478.32 & 416.46 & 0.5593 \\
\hline
\end{tabular}
\caption{Effect of different random seeds on model performance, evaluated over 200 episodes, 1000 steps per episode and window size 128$\times$128}
\label{tab:random_seed_results}
\end{table}

\section{Results}

\begin{table*}[h]
\centering
\begin{tabular}{|c|c|c|c|}
\hline
\textbf{Model} & \textbf{OCR F1 Score ↑} & \textbf{OCR Accuracy ↑} & \textbf{CLIPScore ↑} \\
\hline
ControlNet \cite{controlnet}           & 58.65 &  23.90 & 34.24 \\
DeepFloyd \cite{deepfloyd}            & 17.62 & 2.620  & 32.67 \\
PixArt-$\alpha$ \cite{pixart}         & 0.03 & 0.02 & 27.88 \\
GlyphControl \cite{glyphcontrol}       & 64.07 & 32.56 & 34.56 \\
TextDiffuser \cite{textdiffuser}        & 78.24 & 56.09 & 34.36 \\
TextDiffuser-2 \cite{textdiffuser2}      & 75.06 & 57.58 & 34.50 \\
TextDiffuser-RL (Ours)                  & \textbf{69.91} & \textbf{34.30} & \textbf{34.47} \\
\hline
\end{tabular}
\caption{Comparison of text-to-image models on the MARIOEval benchmark. OCR F1 and Accuracy were computed using the PaddleOCR PP-OCRv5 \cite{paddleocr} model. CLIPScore was measured using the CLIP ViT-B/16 \cite{clip}.}
\label{tab:model_comparison}
\end{table*}

\subsection{Evaluation and Comparison}
We evaluated the performance of different SOTA text layout generation models, evaluated the visual quality of generated images in this section. We also compared the inference efficiency of our approach.

\subsubsection{Text-to-Image Generation Quality Analysis}
We used CLIPScore~\cite{clip} and OCR metrics~\cite{graves2006connectionist} to evaluate the visual and textual quality of the generated images. CLIPScore measures the semantic similarity between the input text prompt and the generated image by comparing their embeddings using the CLIP model. This metric helps assess how well the image reflects the intended textual content. However, the OCR F1 score evaluates how accurately the generated text can be extracted from the image using an optical character recognition (OCR) engine. In addition, we used the MARIOEval~\cite{textdiffuser} benchmark to measure these metrics. The dataset consists of 5,414 image-description pairs designed to evaluate the quality of multimodal generation.

As shown in Table~\ref{tab:model_comparison}, TextDiffuser and TextDiffuser-2 exhibit strong performance in OCR and CLIPScore, benefiting from LM-based layout generation. While our RL-based approach does not rely on such heavy architectures, it still achieves competitive OCR F1 (69.91) and CLIPScore (34.47). It demonstrates the effectiveness of reinforcement learning for layout optimization. Our model achieves nearly comparable accuracy with drastically lower resource demands. For example, compared to ControlNet and DeepFloyd, which have OCR F1 scores of 58.65 and 17.62 respectively, TextDiffuser-RL performs much better while using a simpler method. Even GlyphControl, with an OCR F1 of 64.07, is slightly below our model. This makes TextDiffuser-RL an efficient alternative without a significant performance compromise.

\subsubsection{Inference Time Comparison}
We compared the time efficiency and resource requirements of the language model M1 from TextDiffuser with our RL-based bounding box optimization. The RL models run significantly faster, taking approximately 60 milliseconds on CPU and 70 milliseconds on GPU per inference. In contrast, the M1 model requires around 2960 milliseconds on GPU and often fails to load on CPU due to high memory demands. This evaluation was conducted on Google Colab using an Intel Xeon CPU (2.20 GHz) and an NVIDIA L4 GPU. The L4 GPU features 24 GB of GDDR6 memory and is optimized for low-latency AI inference and also offering high throughput for tasks such as image generation and text-to-image diffusion. This evaluation was conducted on the LAIONEVal4000~\cite{schuhmann2021laion400m} dataset, a subset of MARIOEval containing 4,000 diverse prompt-image pairs.

\begin{table}[h!]
\centering
\resizebox{\columnwidth}{!}{%
\begin{tabular}{|c|c|c|c|}
\hline
\textbf{Model} & \textbf{CPU Time (ms)} & \textbf{GPU Time (ms)} & \textbf{Memory Usage} \\
\hline
PPO & \textbf{60} & \textbf{70} & \textbf{$\sim$2 MB RAM} \\
M1 model & N/A & 2960 & 13.477 GB VRAM \\
\hline
\end{tabular}}
\caption{Time and memory comparison between PPO and M1~\cite{textdiffuser2} for bounding box optimization, averaged over LAIONEval4000 prompts.}
\label{tab:time-comparison}
\end{table}

Additionally, the RL approach is far more memory efficient, consuming approximately 2~MB of system RAM. On the other hand, the M1 language model requires 14–16~GB of GPU memory (VRAM) to generate images. This dramatic reduction in both runtime and memory usage makes the RL-based method a highly practical alternative for bounding box layout optimization.

\subsection{Generated Results}
To qualitatively evaluate our method, we visualize both the final generated images and the corresponding PPO-optimized bounding box layouts used as spatial priors.

\begin{figure}[H]
    \centering
    % Row 1
    \includegraphics[width=0.23\linewidth, height=2.5cm]{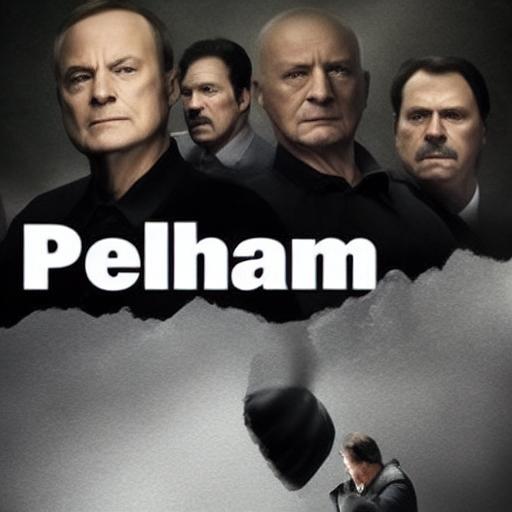}
    \includegraphics[width=0.23\linewidth, height=2.5cm]{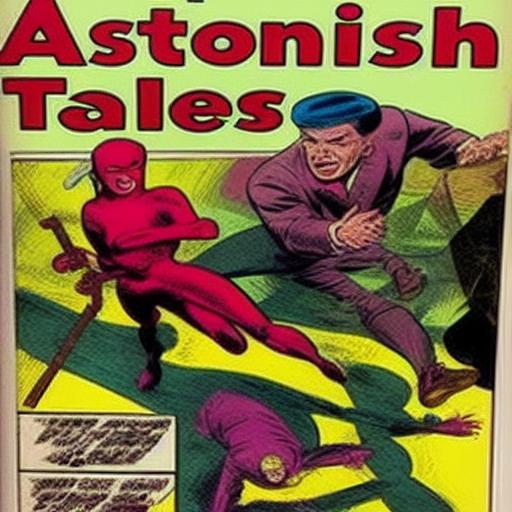}
    \includegraphics[width=0.23\linewidth, height=2.5cm]{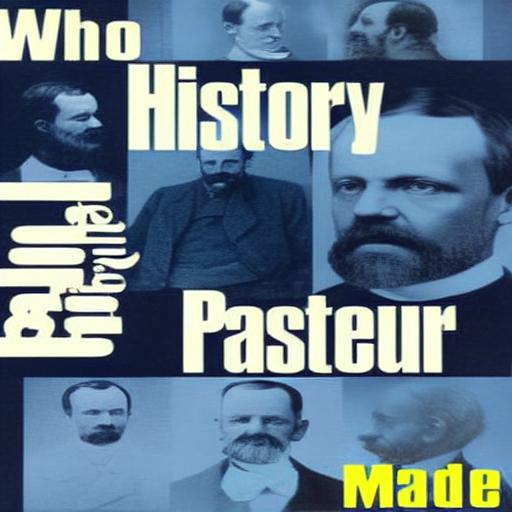}
    \includegraphics[width=0.23\linewidth, height=2.5cm]{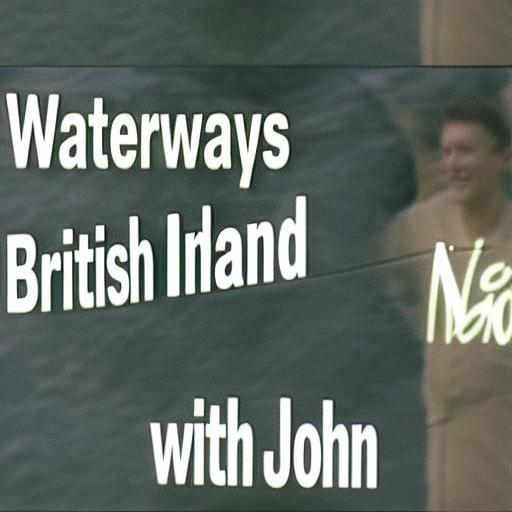}

    \vspace{0.5em}
    % Row 2
    \includegraphics[width=0.23\linewidth, height=2.5cm]{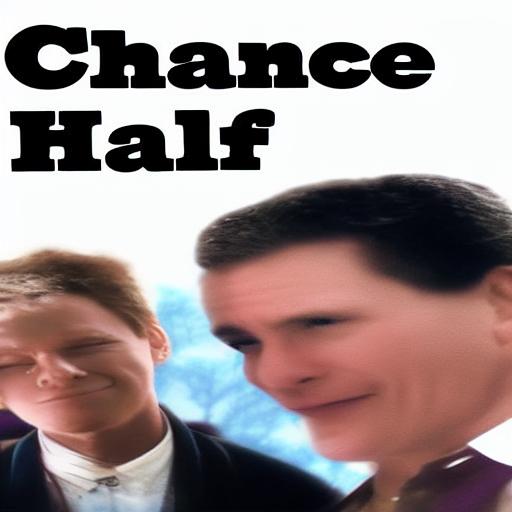}
    \includegraphics[width=0.23\linewidth, height=2.5cm]{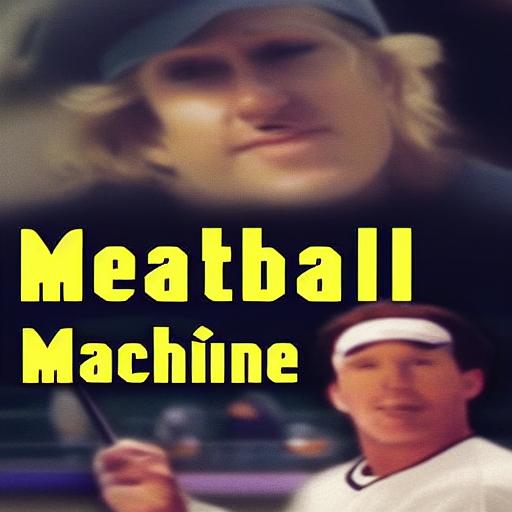}
    \includegraphics[width=0.23\linewidth, height=2.5cm]{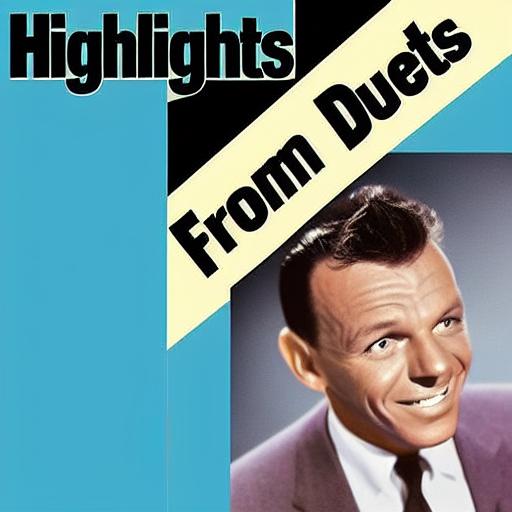}
    \includegraphics[width=0.23\linewidth, height=2.5cm]{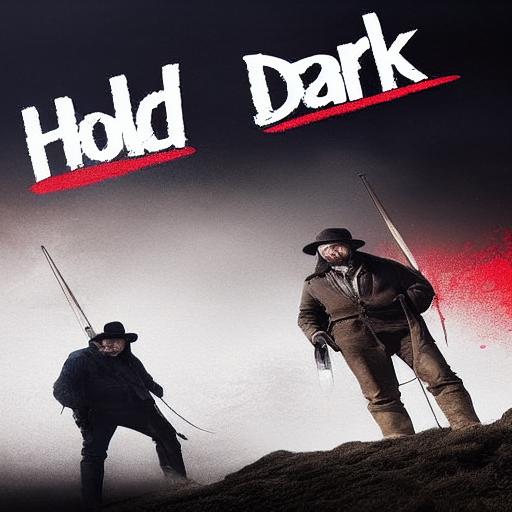}

    \vspace{0.5em}
    % Row 3
    \includegraphics[width=0.23\linewidth, height=2.5cm]{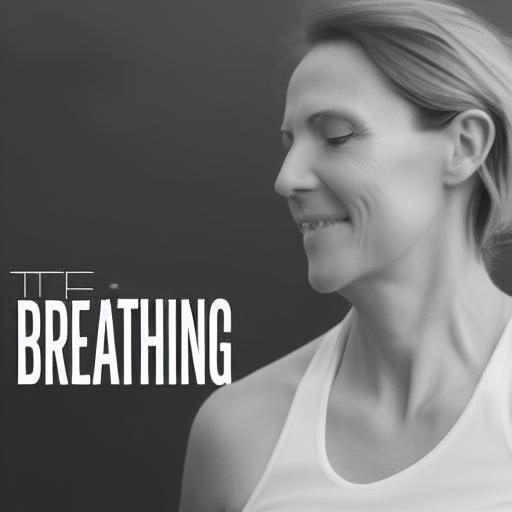}
    \includegraphics[width=0.23\linewidth, height=2.5cm]{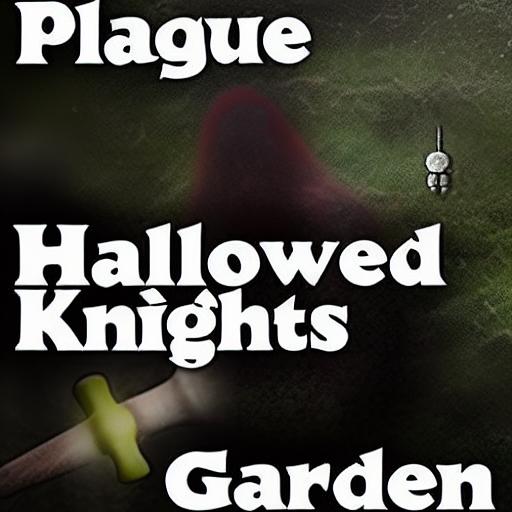}
    \includegraphics[width=0.23\linewidth, height=2.5cm]{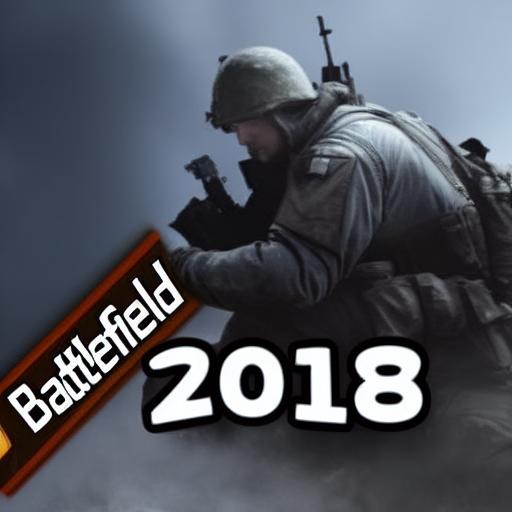}
    \includegraphics[width=0.23\linewidth, height=2.5cm]{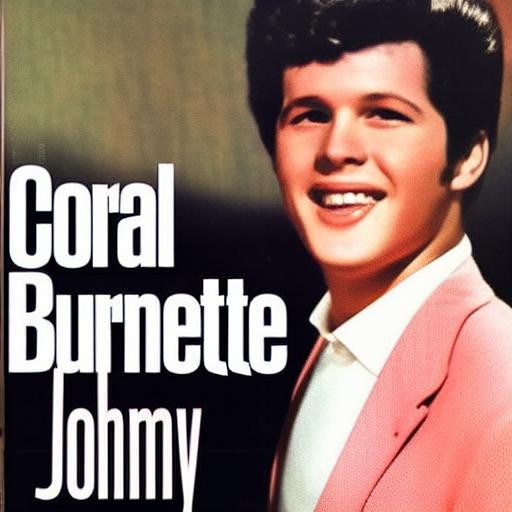}

    \caption{TextDiffuser-RL results on prompts like book covers and posters}
    \label{fig:generated_images}
\end{figure}

\begin{figure}[H]
    \centering
    % Row 1
    \includegraphics[width=0.23\linewidth]{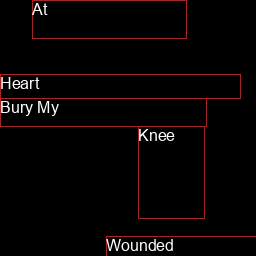}
    \includegraphics[width=0.23\linewidth]{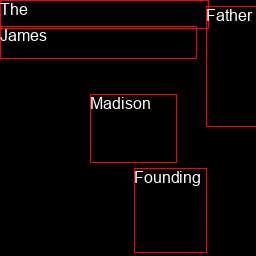}
    \includegraphics[width=0.23\linewidth]{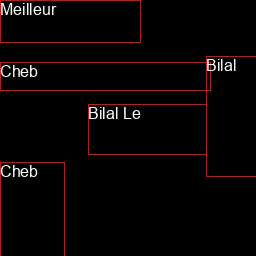}
    \includegraphics[width=0.23\linewidth]{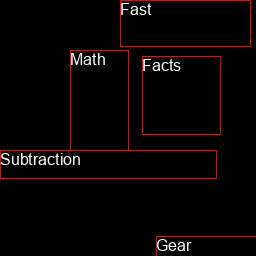}

    \vspace{0.5em}
    % Row 2
    \includegraphics[width=0.23\linewidth]{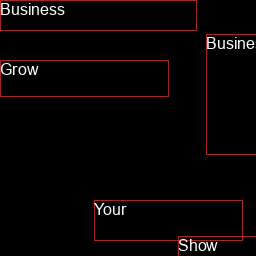}
    \includegraphics[width=0.23\linewidth]{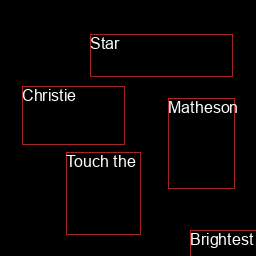}
    \includegraphics[width=0.23\linewidth]{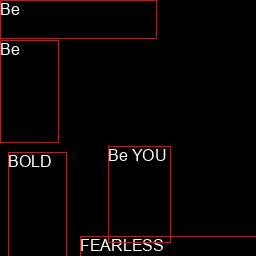}
    \includegraphics[width=0.23\linewidth]{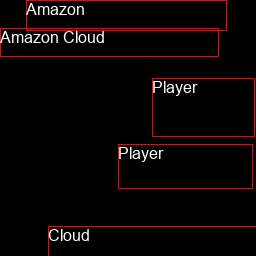}
    
    \vspace{0.5em}
    % Row 3
    \includegraphics[width=0.23\linewidth]{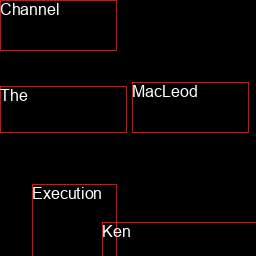}
    \includegraphics[width=0.23\linewidth]{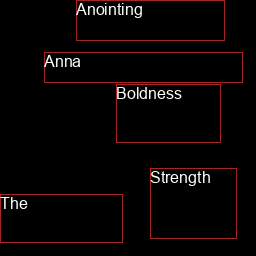}
    \includegraphics[width=0.23\linewidth]{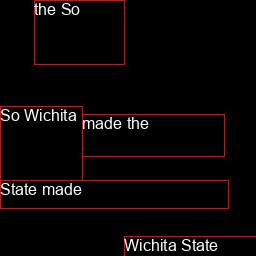}
    \includegraphics[width=0.23\linewidth]{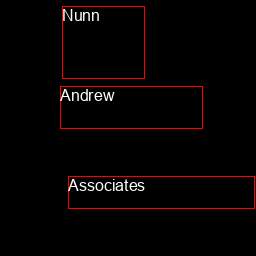}

    \caption{PPO-trained layouts from GlyphEnv with minimal text overlap}
    \label{fig:generated_box}
\end{figure}

\section{Discussion}

Our experimental results demonstrate that reinforcement learning—particularly PPO—offers a compelling approach for bounding box layout generation, significantly reducing overlap while maintaining high inference efficiency.

Our method has several limitations, despite these strengths. First, the current environment is static: it generates a fixed number of bounding boxes and assumes uniform box sizes, irrespective of prompt length or spatial context. This restricts adaptability in handling prompts with varying complexity. Additionally, in edge cases involving dense or unusually long text prompts, the model may still produce overlapping boxes. The system also lacks enforcement of sentence-level or semantic ordering, which can affect coherence in generated layouts.

Moreover, while PPO effectively optimizes for non-overlap and spatial separation, it does not explicitly account for human-aligned positioning conventions—such as placing titles at the top or centering key content. As a result, the aesthetic quality and functional flow of the layout may degrade, particularly in scenarios requiring structured or hierarchically organized text.

Future work may explore dynamic environments that support variable numbers and sizes of bounding boxes, potentially conditioned on the linguistic structure of the prompt. Integrating PPO with heuristic post-processing or learned priors could further enhance robustness. Other promising directions include incorporating character-aware reward functions, enabling non-rectangular text regions, and user-controllable layout tools to improve practical usability in real-world design workflows.

\section{Conclusion}
We present \textbf{TextDiffuser-RL}, a reinforcement learning-based approach that enhances text-to-image generation by optimizing text layouts without relying on large language models. By integrating our custom environment, \texttt{GlyphEnv}, with a diffusion-based generation backbone, the system learns to place text in a visually coherent, non-overlapping manner—achieving competitive image quality and semantic alignment at a fraction of the computational cost. Our results demonstrate that reinforcement learning offers a practical, lightweight alternative to traditional layout planning methods, enabling high-quality text-to-image synthesis even on resource-constrained devices. While challenges remain in handling particularly dense or complex prompts, this work establishes a foundation for more efficient and adaptive layout systems. TextDiffuser-RL sets a new benchmark for efficient and intelligent visual synthesis. Future directions include support for multilingual prompts, user-guided layout control, and further integration with content-aware or structure-aware generation models—broadening the system’s applicability in real-world design and accessibility applications.

\newpage

\bibliographystyle{unsrt}

\begin{thebibliography}{00}

\bibitem{zhao2023udifftext}
Y. Zhao and Z. Lian, ``UDiffText: A Unified Framework for High-quality Text Synthesis in Arbitrary Images via Character-aware Diffusion Models,'' 
arXiv preprint arXiv:2312.04884, 2023. Available: \url{https://arxiv.org/abs/2312.04884}.

\bibitem{liu2024glyphbyt5}
Z. Liu, W. Liang, Z. Liang, C. Luo, J. Li, G. Huang, and Y. Yuan, ``Glyph-ByT5: A Customized Text Encoder for Accurate Visual Text Rendering,'' 
arXiv preprint arXiv:2403.09622, 2024. Available: \url{https://arxiv.org/abs/2403.09622}.

\bibitem{liu2023characteraware}
R. Liu, D. Garrette, C. Saharia, W. Chan, A. Roberts, S. Narang, I. Blok, R. J. Mical, M. Norouzi, and N. Constant, 
``Character-Aware Models Improve Visual Text Rendering,'' 
arXiv preprint arXiv:2212.10562, 2023. Available: \url{https://arxiv.org/abs/2212.10562}.

\bibitem{chen2024enhancing}
C. Chen, A. Wang, H. Wu, L. Liao, W. Sun, Q. Yan, and W. Lin, 
``Enhancing Diffusion Models with Text-Encoder Reinforcement Learning,'' 
arXiv preprint arXiv:2311.15657, 2024. Available: \url{https://arxiv.org/abs/2311.15657}.

\bibitem{wang2024dreamtext}
Y. Wang, W. Zhang, and C. Jin, 
``DreamText: High Fidelity Scene Text Synthesis,'' 
arXiv preprint arXiv:2405.14701, 2024. Available: \url{https://arxiv.org/abs/2405.14701}.

\bibitem{wei2024powerful}
F. Wei, W. Zeng, Z. Li, D. Yin, L. Duan, and W. Li, 
``Powerful and Flexible: Personalized Text-to-Image Generation via Reinforcement Learning,'' 
arXiv preprint arXiv:2407.06642, 2024. Available: \url{https://arxiv.org/abs/2407.06642}.

\bibitem{miao2024subjectdriven}
Y. Miao, W. Loh, S. Kothawade, P. Poupart, A. Rashwan, and Y. Li, 
``Subject-driven Text-to-Image Generation via Preference-based Reinforcement Learning,'' 
arXiv preprint arXiv:2407.12164, 2024. Available: \url{https://arxiv.org/abs/2407.12164}.

\bibitem{zhang2024large}
Z. Zhang, Y. Wang, J. Liu, and R. Liu, 
``Large-scale Reinforcement Learning for Diffusion Models,'' 
arXiv preprint arXiv:2401.12244, 2024. Available: \url{https://arxiv.org/abs/2401.12244}.

\bibitem{schulman2017proximal}
J. Schulman, F. Wolski, P. Dhariwal, A. Radford, and O. Klimov, 
``Proximal Policy Optimization Algorithms,'' 
arXiv preprint arXiv:1707.06347, 2017. Available: \url{https://arxiv.org/abs/1707.06347}.

\bibitem{zhang2023dpok}
Y. Zhang, J. Wong, L. Chen, and R. Liu, 
``DPOK: Reinforcement Learning for Fine-tuning Text-to-Image Diffusion Models,'' 
arXiv preprint arXiv:2305.16381, 2023. Available: \url{https://arxiv.org/abs/2305.16381}.

\bibitem{zhang2023robotics}
Y. Zhang, W. Chen, S. Kothawade, P. Poupart, and A. Rashwan, 
``Can Pre-Trained Text-to-Image Models Generate Visual Goals for Robotics?,'' 
in \textit{Advances in Neural Information Processing Systems (NeurIPS)}, New Orleans, LA, 2023.

\bibitem{ho2020denoising}
J. Ho, A. Jain, and P. Abbeel, 
``Denoising Diffusion Probabilistic Models,'' 
in \textit{Advances in Neural Information Processing Systems (NeurIPS)}, vol. 33, pp. 6840--6851, Virtual, 2020.

\bibitem{saharia2022imagen}
C. Saharia, W. Chan, S. Saxena, J. Li, P. R. Y. Roberts, and I. Blok, 
``Imagen: Text-to-Image Diffusion Models Beat GANs on FID and CLIP Scores,'' 
in \textit{Proceedings of the 39th International Conference on Machine Learning (ICML)}, vol. 162, pp. 9187--9198, Baltimore, MD, 2022.

\bibitem{rombach2022high}
R. Rombach, A. Blattmann, D. Esser, and B. Ommer, 
``High-Resolution Text-to-Image Synthesis with Latent Diffusion Models,'' 
in \textit{Proceedings of the IEEE/CVF Conference on Computer Vision and Pattern Recognition (CVPR)}, pp. 10684--10694, New Orleans, LA, 2022.

\bibitem{nichol2021guided}
A. Nichol, J. Dhariwal, D. Li, D. K. Jain, and A. K. Jain, 
``Guided Diffusion Models for Text-Driven Image Manipulation,'' 
in \textit{Proceedings of the 38th International Conference on Machine Learning (ICML)}, vol. 139, pp. 16469--16480, Virtual, 2021.

\bibitem{ouyang2023reinforcement}
D. Ouyang, C. Zhang, Y. Wu, J. Wang, and I. Blok, 
``Reinforcement Learning with Human Feedback for Generative Models,'' 
arXiv preprint arXiv:2302.03426, 2023. Available: \url{https://arxiv.org/abs/2302.03426}.

\bibitem{chen2023text}
Z. Chen, H. Zhang, and R. Liu, 
``Text Rendering with Neural Diffusion Models,'' 
arXiv preprint arXiv:2304.01587, 2023. Available: \url{https://arxiv.org/abs/2304.01587}.

\bibitem{paddleocr}
Y. Du, X. Bai, T. He, \textit{et al.}, 
``PP-OCR: A Practical Ultra Lightweight OCR System,'' 
arXiv preprint arXiv:2009.09941, 2021. Available: \url{https://arxiv.org/abs/2009.09941}.

\bibitem{clip}
A. Radford, J. W. Kim, C. Hallacy, \textit{et al.}, 
``Learning Transferable Visual Models From Natural Language Supervision,'' 
in \textit{Proceedings of the 38th International Conference on Machine Learning (ICML)}, Virtual, 2021.

\bibitem{ddpg}
T. P. Lillicrap, J. J. Hunt, A. Pritzel, N. Heess, T. Erez, Y. Tassa, D. Silver, and D. Wierstra, 
``Continuous Control with Deep Reinforcement Learning,'' 
arXiv preprint arXiv:1509.02971, 2015. Available: \url{https://arxiv.org/abs/1509.02971}.

\bibitem{haarnoja2018soft}
T. Haarnoja, A. Zhou, P. Abbeel, and S. Levine, 
``Soft Actor-Critic: Off-Policy Maximum Entropy Deep Reinforcement Learning with a Stochastic Actor,'' 
arXiv preprint arXiv:1801.01290, 2018. Available: \url{https://arxiv.org/abs/1801.01290}.

\bibitem{textdiffuser}
B. Zhou, Y. Ji, J. Han, M. Liu, Q. Jin, and P. Zhou, 
``TextDiffuser: Diffusion Models for Text Rendering and Synthesis,'' 
in \textit{Proceedings of the IEEE/CVF Conference on Computer Vision and Pattern Recognition (CVPR)}, pp. 6706--6715, Vancouver, Canada, 2023.

\bibitem{saocrpaint}
H. Chen, T. Wang, and J. Zhu, 
``SA-OcrPaint: Layout-aware OCR-guided Inpainting for Text-Embedded Image Synthesis,'' 
in \textit{Proceedings of the IEEE/CVF Conference on Computer Vision and Pattern Recognition (CVPR)}, Vancouver, Canada, 2023.

\bibitem{artist2023}
K. Lin, H. Zhou, J. Huang, Z. Li, and J. Wang, 
``ARTIST: Disentangled Text-to-Image Generation with Text Structure Guidance,'' 
in \textit{Proceedings of the IEEE/CVF Conference on Computer Vision and Pattern Recognition (CVPR)}, Vancouver, Canada, 2023. Available: \url{https://arxiv.org/html/2406.12044v2}.

\bibitem{jaccard1901}
P. Jaccard, 
``Étude comparative de la distribution florale dans une portion des Alpes et des Jura,'' 
\textit{Bulletin de la Société Vaudoise des Sciences Naturelles}, vol. 37, pp. 547--579, 1901.

\bibitem{graves2006connectionist}
A. Graves, S. Fernandez, F. Gomez, and J. Schmidhuber, 
``Connectionist Temporal Classification: Labelling Unsegmented Sequence Data with Recurrent Neural Networks,'' 
in \textit{Proceedings of the 23rd International Conference on Machine Learning (ICML)}, pp. 369--376, Pittsburgh, PA, 2006.

\bibitem{schuhmann2021laion400m}
C. Schuhmann, R. Beaumont, R. Vencu, C. Gordon, T. Kaczmarczyk, T. Coombes, C. Schmid, and D. Kiela, 
``LAION-400M: Open Dataset of CLIP-Filtered 400 Million Image-Text Pairs,'' 
arXiv preprint arXiv:2111.02114, 2021. Available: \url{https://arxiv.org/abs/2111.02114}.

\bibitem{textdiffuser2}
J. Chen, Y. Huang, T. Lv, L. Cui, Q. Chen, and F. Wei, 
``TextDiffuser-2: Unleashing the Power of Language Models for Text Rendering,'' 
arXiv preprint arXiv:2311.16465, 2023. Available: \url{https://arxiv.org/abs/2311.16465}.

\bibitem{glyphcontrol}
Y. Yang, D. Gui, Y. Yuan, W. Liang, H. Ding, H. Hu, and K. Chen, 
``GlyphControl: Glyph Conditional Control for Visual Text Generation,'' 
arXiv preprint arXiv:2305.18259, 2023. Available: \url{https://arxiv.org/abs/2305.18259}.

\bibitem{pixart}
J. Chen, J. Yu, C. Ge, L. Yao, E. Xie, Y. Wu, Z. Wang, J. Kwok, P. Luo, H. Lu, and Z. Li, 
``PixArt-$\alpha$: Fast Training of Diffusion Transformer for Photorealistic Text-to-Image Synthesis,'' 
arXiv preprint arXiv:2310.00426, 2023. Available: \url{https://arxiv.org/abs/2310.00426}.

\bibitem{controlnet}
L. Zhang, A. Rao, and M. Agrawala, 
``Adding Conditional Control to Text-to-Image Diffusion Models,'' 
arXiv preprint arXiv:2302.05543, 2023. Available: \url{https://arxiv.org/abs/2302.05543}.

\bibitem{deepfloyd}
C. Saharia, W. Chan, S. Saxena, L. Li, J. Whang, E. Denton, S. K. S. Ghasemipour, B. K. Ayan, S. S. M. Mahdavi, R. G. Lopes, T. Salimans, J. Ho, D. J. Fleet, and M. Norouzi, 
``Photorealistic Text-to-Image Diffusion Models with Deep Language Understanding,'' 
arXiv preprint arXiv:2205.11487, 2022. Available: \url{https://arxiv.org/abs/2205.11487}.


\end{thebibliography}

\newpage
\section{Appendix}

\subsection{Hyper-parameter Selection}
Specifically, we used a learning rate of 0.0003, a rollout size of 2048, a batch size of 64, and ran 10 training epochs per update. These are commonly used PPO values and perform reliably in our environment. We set the discount factor $\gamma$ to 0.99 and used Generalized Advantage Estimation (GAE) with $\lambda = 0.95$ to manage the bias-variance trade-off effectively. We applied a clip range of 0.2 and disabled entropy regularization for optimization. We used a value function loss coefficient of 0.5, applied gradient clipping with a maximum norm of 0.5, and enabled advantage normalization. These settings helped ensure stable and consistent training. All experiments were monitored using TensorBoard. This configuration closely follows the standard and widely adopted hyperparameter setup for PPO. It was also established in the original OpenAI baseline and many subsequent implementations.

\subsection{Prompt Preprocessing}
Prompts pre-processed using Python's \texttt{re} module to extract quoted phrases using simple regular expression. These phrases are split into words, then joined with slashes, then used as keywords for image generation.

\subsection{Training with different numbers of rectangular \& area}
\begin{figure}[h]
    \centering
    \includegraphics[width=0.9\linewidth, height=4cm]{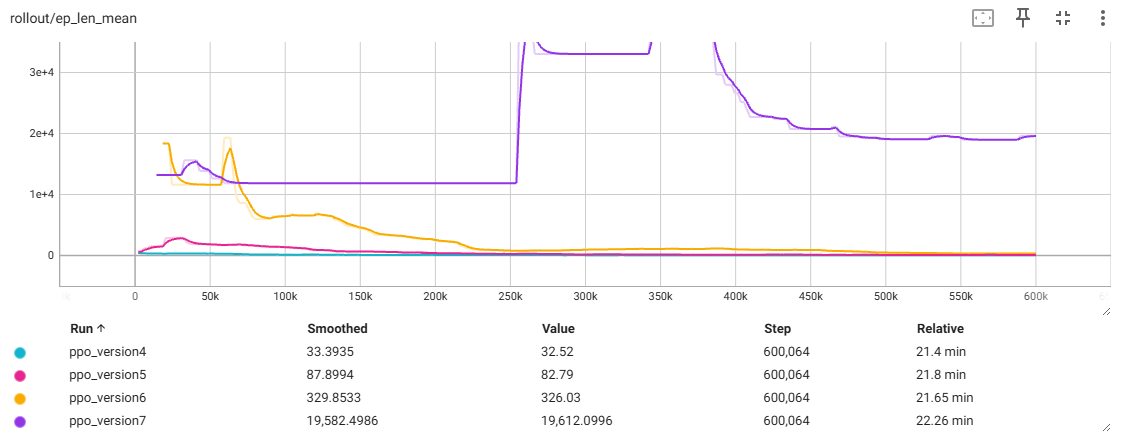}
    \caption{Mean episode length over training steps for different numbers of rectangles.}
    \label{fig:ep_len_mean_1}
\end{figure}

\begin{figure}[h]
    \centering
    \includegraphics[width=0.9\linewidth, height=4cm]{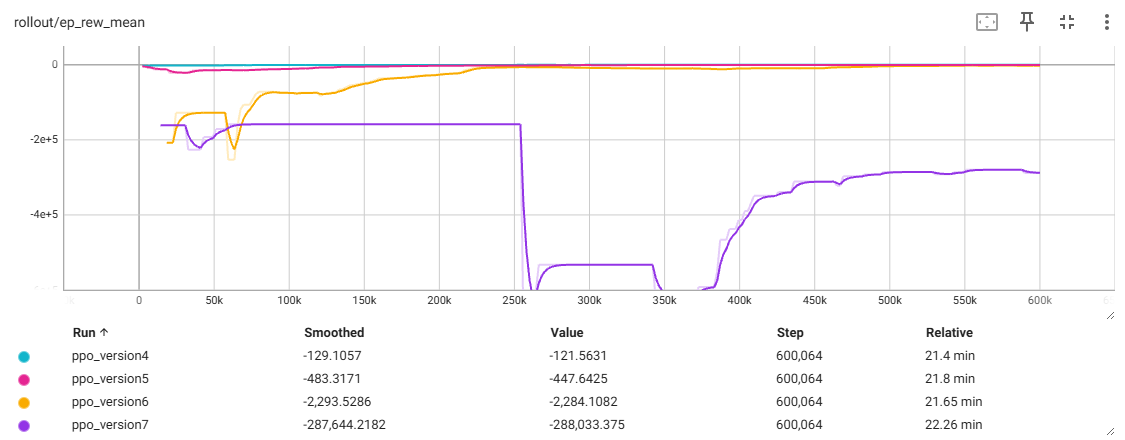}
    \caption{Mean episode reward over training steps for different numbers of rectangles.}
    \label{fig:ep_rew_mean_1}
\end{figure}

We trained PPO agents in environments with 4, 5, 6, and 7 rectangular objects—corresponding to \texttt{ppo\_version4}, \texttt{ppo\_version5}, \texttt{ppo\_version6}, and \texttt{ppo\_version7}, respectively. As shown in Figure~\ref{fig:ep_len_mean_1} and Figure~\ref{fig:ep_rew_mean_1}, the agent learned fastest when there were fewer rectangles. For example, with 4 rectangles (\texttt{ppo\_version4}), the agent quickly achieved short episode lengths and reasonable rewards. As the number of rectangles increased, learning became harder. With 7 rectangles (\texttt{ppo\_version7}), the agent’s training was unstable—episode lengths were very high at first, and rewards dropped sharply. Overall, more rectangular objects made the task more complex, slowing down learning and reducing performance.

Again, we trained PPO agents with the same number of rectangular regions but varied the total area sizes like 1300, 1500, 1800, and 2000 units. These configurations are labeled as \texttt{ppo\_version8}, \texttt{ppo\_version9}, \texttt{ppo\_version11}, and \texttt{ppo\_version10}, respectively.

\begin{figure}[h]
    \centering
    \includegraphics[width=0.9\linewidth, height=4cm]{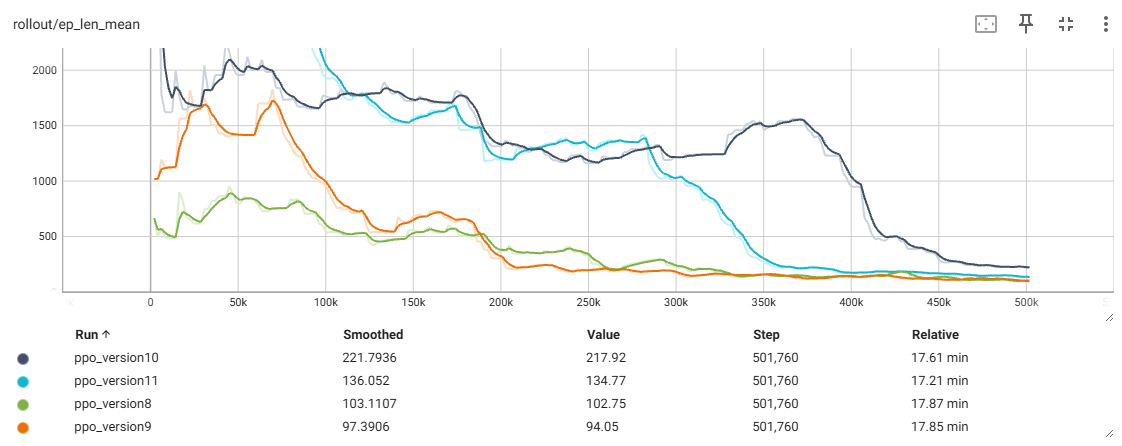}
    \caption{Mean episode length over training steps for different numbers of rectangles.}
    \label{fig:ep_len_mean_2}
\end{figure}

From the Figure~\ref{fig:ep_len_mean_2}, we can see that agents trained on larger areas tend to take longer to complete an episode in the early stages. However, over time, all versions converge towards shorter episode lengths, with \texttt{ppo\_version8 \& ppo\_version9} stabilizing earlier. The Figure~\ref{fig:ep_len_mean_2} plot shows a clear improvement trend across all area settings. Smaller area configurations \texttt{ppo\_version8 \& ppo\_version9} reach higher average rewards faster, while larger areas \texttt{ppo\_version10 \& ppo\_version11} take more steps to improve. Eventually, all versions approach similar performance, but with different learning speeds. These results suggest that larger areas introduce more complexity, while smaller areas allow faster convergence.

\begin{figure}[h]
    \centering
    \includegraphics[width=0.9\linewidth, height=4cm]{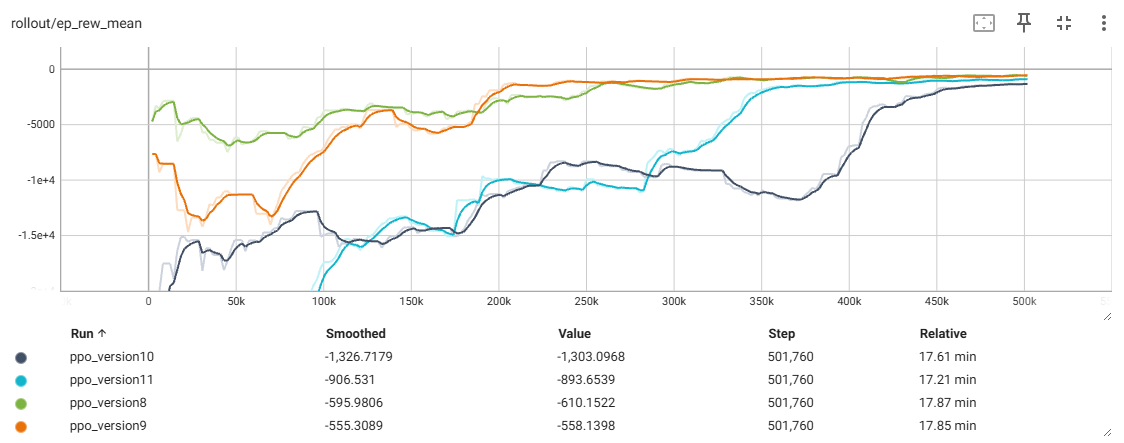}
    \caption{Mean episode reward over training steps for different numbers of rectangles.}
    \label{fig:ep_rew_mean_2}
\end{figure}

\subsection{Random seed comparison}
\begin{figure}[h]
    \centering
    \includegraphics[width=0.9\linewidth, height=4cm]{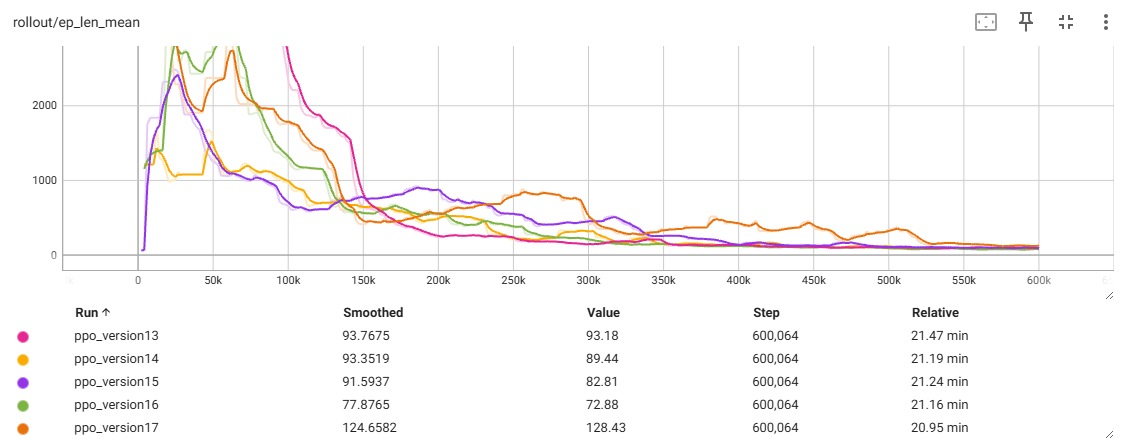}
    \caption{Mean episode length over training steps for different numbers of rectangles.}
    \label{fig:ep_len_mean_3}
\end{figure}

\begin{figure}[h]
    \centering
    \includegraphics[width=0.9\linewidth, height=4cm]{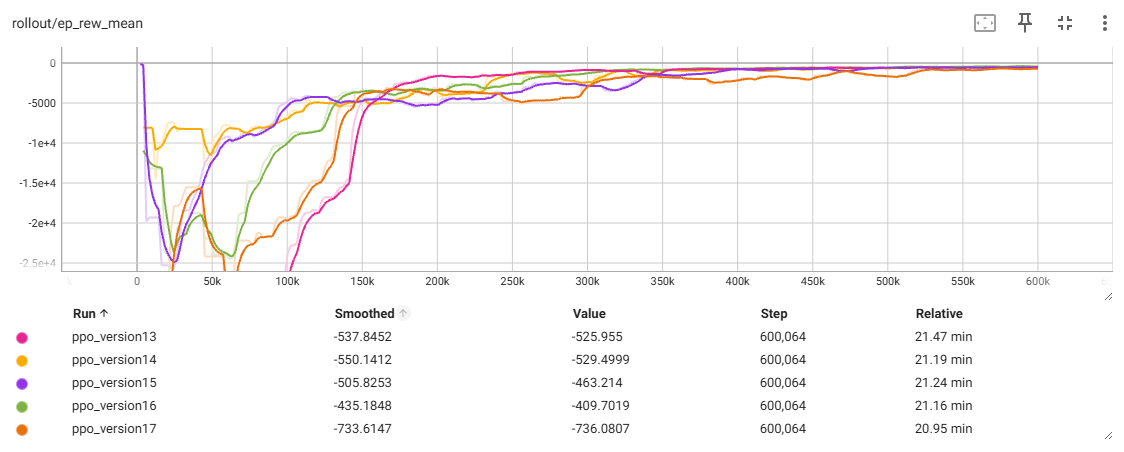}
    \caption{Mean episode reward over training steps for different numbers of rectangles.}
    \label{fig:ep_rew_mean_3}
\end{figure}

To understand how random seed affects the performance of PPO in our custom environment, we trained the model using five different seeds, like 0, 42, 123, 551, and 999. These runs are labeled as \texttt{ppo\_version13} to \texttt{ppo\_version17}. The results are shown in Figure~\ref{fig:ep_len_mean_3} and Figure~\ref{fig:ep_rew_mean_3}. Although there were some variations in the early stages of training—such as differences in the peak episode lengths or the rate of reward improvement—all models eventually converged to similar performance levels in terms of both episode length and average reward. This indicates that the PPO algorithm is stable across different random seeds in our custom environment. The outcomes show only minor differences, which suggests that randomness introduced by different seeds does not significantly impact the model's learning process or its final performance.

\subsection{Environmental Setup}
All experiments were run on Google Colab, which utilizes its free GPU resources for faster training. The gymnasium was used to create and manage custom environments. Stable-Baselines3 provided reliable implementations of RL algorithms to speed up training. Transformers enabled easy use of powerful language models. PaddleOCR was included to handle any text recognition tasks from images. TensorBoard helped monitor training performance in real time.

\subsection{Hardware Specification}
The experiments were conducted on Google Colab, which runs a Linux-based Ubuntu environment. The CPU provided was a virtualized Intel Xeon processor, suitable for general computations. For accelerated training, two types of GPUs were used: the NVIDIA Tesla T4 and NVIDIA L4. The NVIDIA Tesla T4 GPU features 2560 CUDA cores, 16 GB of GDDR6 memory, and a boost clock of around 1590 MHz. The NVIDIA L4 GPU is a more recent model, equipped with 3072 CUDA cores, 8 GB of GDDR6 memory.

\end{document}